\documentclass[conference]{IEEEtran}
\IEEEoverridecommandlockouts
\usepackage{cite}
\usepackage{amsmath,amssymb,amsfonts}
\usepackage{algorithmic}
\usepackage{graphicx}
\usepackage{textcomp}
\usepackage{xcolor}

\usepackage{hyperref}

\def\BibTeX{{\rm B\kern-.05em{\sc i\kern-.025em b}\kern-.08em
    T\kern-.1667em\lower.7ex\hbox{E}\kern-.125emX}}
\begin{document}

\title{Higgs Boson Classification: Brain-inspired BCPNN Learning with \emph{StreamBrain}}

\author{\IEEEauthorblockN{Martin Svedin, Artur Podobas, Steven W. D. Chien, Stefano Markidis}
\IEEEauthorblockA{
\textit{School of Electrical Engineering and Computer Science}\\
\textit{KTH Royal Institute of Technology}\\
Stockholm, Sweden \\
}
}

\maketitle

\begin{abstract}
One of the most promising approaches for data analysis and exploration of large data sets is Machine Learning (ML) techniques that are inspired by brain models. Such methods use alternative learning rules potentially more efficiently than established learning rules. In this work, we focus on the potential of brain-inspired ML for exploiting High-Performance Computing (HPC) resources to solve ML problems: we discuss the BCPNN and an HPC implementation, called StreamBrain, its computational cost, suitability to HPC systems. As an example, we use StreamBrain to analyze the Higgs Boson dataset from High Energy Physics and discriminate between background and signal classes in collisions of high-energy particle colliders. Overall, we reach up to 69.15\% accuracy and 76.4\% Area Under the Curve (AUC) performance.
\end{abstract}


\section{Introduction}
Today, Machine Learning (ML) as a method for data exploration and understanding has permeated nearly all scientific fields and disciplines. Many of these are so-called \textit{Deep Neural Networks} (DNNs) and are based on backpropagation~\cite{lecun2015deep}, which has been motivated by the super-human level of performance they can achieve in image recognition tasks~\cite{shankar2020evaluating}. Recently, however, the Bayesian Confidence Propagation Neural Network (BCPNN)~\cite{ravichandran2020learning} has emerged as an alternative to Deep Learning (DL). BCPNN – unlike DL – is brain-inspired and has an architecture that closely mimics that of the human brain. The basic units for computation are hyper- and mini-columns, which are structures to be fundamental in the human neocortex~\cite{mountcastle1997columnar}. Furthermore, BCPNN feature \textit{structural plasticity}—the network does not only learn to interpret what it is seeing (the receptive field) but also learns \textit{where} it should look to extract most of the information. Structural plasticity is particularly important, as it can reveal new knowledge about the data-set (by inspecting the receptive fields) as well as give opportunity for sparse representations—a theme that is becoming more and more important overall in ML. Finally, BCPNN supports supervised, semi-supervised, and – perhaps most importantly – unsupervised forms of training, which allows bringing order even to unlabeled (the majority) of data.

Today, BCPNN is capable of reaching up to 98.6+\% of testing accuracy on the well-known MNIST~\cite{lecun1998mnist} image set. While this performance is lower than those of backpropagation-based machine learning techniques, we believe that different approach of BCPNN allows for a different insight into the properties of data. Furthermore, a recent high-performance implementation of BCPNN called \textit{StreamBrain}~\cite{podobas2021streambrain} was incepted, with support for both general-purpose processors (MPI/OpenMP), Graphics Processing Units (GPUs through CUDA), and even Field-Programmable Gate Arrays (FPGAs, using prior knowledge of High-Level Synthesis~\cite{podobas2016empowering,podobas2017designing}), which can also be executed on large-scale supercomputers. The execution performance of StreamBrain is on par with modern frameworks such as PyTorch~\cite{paszke2019pytorch}, with some networks training faster (e.g., MNIST) and others slower (e.g., STL-10~\cite{coates2011analysis}).

In this work, we aspire to apply and explore high-energy physics Higgs Boson data set using StreamBrain in order to understand what new knowledge about the data-set that BCPNN can reveal. In physics, one way to search for new exotic particles is through high-energy particle collisions, such as is done in CERN. Often, a large amount of data is gathered from these collisions, and the task at hand becomes understanding what part of the data belongs to a particular signal and what part of the data is simply noise – in short, a binary classification problem~\cite{baldi2014searching}.
We claim the following contributions:

\begin{itemize}
\item We port (for the first time) a high-energy particle collision data-set to the BCPNN machine learning model and reveal techniques that can be used to preprocess the data to better suit the BCPNN model, and
\item We empirically evaluate both execution performance and accuracy of our particle classifier using state-of-the-art GPU systems as well as analyze the outcome, including the structural plasticity masks to get better (and more unique) insights into the data stream
\end{itemize}

The remaining paper is structured as follows: Section \ref{background1} gives a deeper introduction to the BCPNN machine learning model and the computational cost for running BCPNN on modern computers. Section~\ref{method} gives an overview of the StreamBrain framework.  In Section~\ref{setup} we explain our experimental setup, which leads to the high-energy particle physics experiment in Section~\ref{result}. We conclude our study with related work and conclusion in Section~\ref{related} and \ref{conclusion} respectively.

\section{Brain-inspired Machine Learning \& BCPNN }
\label{background1}
While many recent neural networks models, such as those based on backpropagation, have shown remarkable – sometimes superhuman – performance at tasks associated with for example image recognition, they still lack many important properties present in the animal brain. Several examples exists~\cite{kumarasinghe2021brain}, including the lack of incremental learning, the inability to evolve, the need for large amounts of labeled data, and the possibility of catastrophic forgetting. These limitations have already been solved by the animal brain, and by imitating (or trying to replicate) the functionality therein, we can possibly come to create even better human-engineered Artificial Intelligence (AI). One such recent model that is brain-inspired is the Bayesian Confidence Propagation Neural Network (BCPNN).

BCPNN is a brain-inspired neural network model that models the problem as a collection of random variables as joint distribution. Each node in the neural network represents a random variable. Edges that connect between the nodes represent the conditional dependence between variables. Variables inside the hidden layer are grouped together to form the so-called Hypercolumn Units (HCUs). Every HCU contains many Minicolumn Units (MCUs). Intuitively, one can consider HCUs to be the learner of features while MCUs are learning the variation of the feature. One example is to recognize features that have been augmented (flipped, rotated, etc.). The BCPNN model supports both spiking- and rate-based models of computation, where the former maps well to neuromorphic hardware while the latter maps well to accelerators.

\subsection{Local Learning}
One major difference between BCPNN and the traditional DL technique is the lack of backpropagation. Rather than using information that communicates backward from the output layer, BCPNN uses a local and brain-like learning rule to update the weights and biases. For this reason, BCPNN's computation model is extremely suitable for parallel training and allows to achieve high scalability. A direct consequence of local learning is that BCPNN allows for unsupervised learning, where the MCUs are free to learn and discover features in an unsupervised manner, before feeding to an output layer that performs classification.

\subsection{BCPNN Computational Cost}
The local-learning nature of BCPNN makes it an attractive alternative learning method on HPC systems. Unlike backpropagation, learning is local and does not need to be communicated. For this reason, one can conceptually launch different BCPNN instances and scale horizontally without the limiting factor on communication. The rate-based BCPNN formulation is also excellently mapped to modern Deep-Learning oriented accelerators. For example, the computation of activation and weight update can be expressed as a GEMM operation that allows using optimized BLAS libraries. We refer to Ref.~\cite{podobas2021streambrain} for the full details.

\subsection{An Intuitive View over BCPNN}
We end this section by presenting a simplified view of the BCPNN model designed for pattern recognition, giving readers an intuitive feeling behind the algorithm ~\cite{ravichandran2020learning}. The BCPNN model for pattern recognition extracts a set of feature representations of the data in a purely unsupervised manner using simple biologically plausible local learning rules. Learning of features in the BCPNN model encompasses two separate processes: one is learning the weights of the connections, and the other process called structural plasticity learns a sparse connectivity structure. Common deep learning models such as CNNs neglect learning the structure of the network and force fixed spatial filters on the data. However, in the human brain, the structure continuously changes, with synapses being formed and removed between neurons. In BCPNN, structural plasticity is a core feature and is used to maximize information extraction from a dataset given a fixed number of connections --  the network learns to look at the most interesting aspects of the input set.

\begin{figure}
	\begin{center}
		\includegraphics[width=0.3\textwidth]{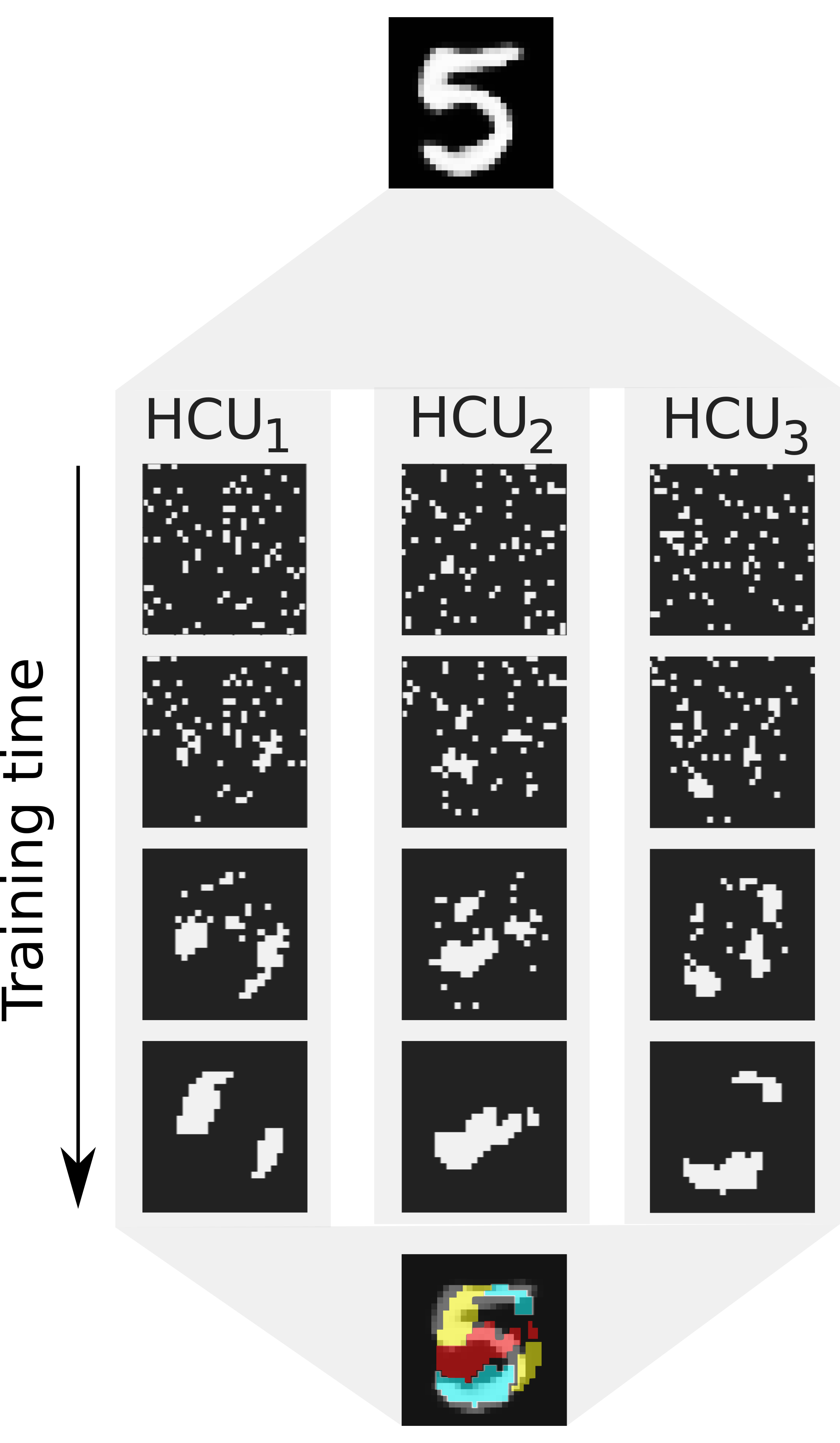}
		\caption{Three different HCUs, which initially looks at random places in the input image dataset (in this example, the number five). Gradually, as we train, the HCUs' learn to look (their \textit{receptive fields}) to extract maximal information from the input image (e.g., the number '5') and ignores patches with little-to-no information at the fringes.}
		\label{fig:structplast}
	\end{center}
	\vspace{-0.5cm}
\end{figure}

To illustrate the property visually, consider the series of images presented in Fig.~\ref{fig:structplast}. Here we see three hypercolumn units (HCUs, the main computational agent in BCPNN), which we are supposed to train so that each HCUs learn a particular feature from the dataset. For simplicity, our input set is MNIST, comprising 28x28 pixel images of handwritten digits. Initially, when we start our training, each HCUs is initiated with a sparse and random receptive field that samples the input image, as represented by the white pixels in the figure. As we begin to train the network, it learns that most of the information about the digits is located in the middle of the images, and each HCU gradually re-evaluates and changes the connectivity (the structure) between itself and the input image. Finally, after several training epochs, the three HCUs have learned precisely which parts of the image to look at to maximize information transfer. We can note that three receptive fields complement each other, and there is little-to-no overlap between the fields. Each HCU comprises several minicolumn units (MCUs), which are trained to offer various interpretations of the HCU's receptive field. These are trained alongside the HCU's, and, as training continues, these are refined to capture different properties from the receptive fields. In this way, BCPNN learns features that have a modular structure, with the HCUs capturing different places in the input, while lower-level MCUs learning diverse features within the HCU’s receptive field.

Finally, unlike existing deep learning techniques, BCPNN is primarily an unsupervised learning technique and uses only supervised learning in the classification layer. As the amount of data grows, methods such as BCPNN can be used to bring order and structure to otherwise seemingly unstructured data.

\section{{\bf StreamBrain}: An HPC Tool for Brain-Inspired Machine Learning}
\label{method}
StreamBrain~\cite{podobas2021streambrain} is a recent framework that implements BCPNN on HPC systems. StreamBrain is primarily described in Python, which is motivated by the portability and the generally increased use of Python within the HPC and ML community~\cite{chien2019tensorflow}. Secondly, since StreamBrain is based on Python, it allows for the integration of BCPNN into the existing Machine Learning pipeline, including the use of hybrid solutions, such as the mixed \textit{BCPNN+Stochastic Gradient Descent (SGD)}-solution present in~\cite{ravichandran2020learning}.

\subsection{High-Performance Compute Backends}
Where-as Python provides portability and ease-of-use, even with the use of BLAS-compatible NUMPY, performance often fails to meet expectations. To remedy the performance limitations of Python, we integrated several hand-coded backends in StreamBrain that targets multiple different technologies: 

\textbf{General-Purpose Processor} support in StreamBrain is provided in two different ways. The first is through a dedicated backend hand-coded in OpenMP and with hand-written SIMD (vector) extensions, which also leverages any available linear algebra library (e.g., Intel MKL~\cite{wang2014intel}) to obtain further performance improvements. The second backend extends the first by providing a Message Passing Interface (MPI) support to facilitate the use of large distributed supercomputers.

\textbf{Graphics Processing Unit (GPU)} support is enabled through a custom Compute Unified Device Architecture (CUDA) implementation. Unlike the general-purpose version above, our GPU version offloads a majority of the BCPNN training loop to this external accelerator in order to minimize any data transfers that would otherwise serve as Amdahl~\cite{amdahl1967validity} serialization points.

\textbf{Field-Programmable Gate Arrays (FPGAs)} is a relatively new accelerator technology in HPC. Traditionally, FPGAs have been used to simulate ASICs prior to tape-out or to accelerate fixed-point precision heavy Digital Signal Processing (DSP) algorithms such as Fourier transform and image processing. Today, however, with the advent of more advanced functionality, FPGA can compete with more general-purpose systems even on the grounds of floating-point performance. In StreamBrain, prototype FPGA support is available using High-Level Synthesis (HLS) through Intel’s OpenCL~\cite{czajkowski2012opencl} framework, and the intention is to honor architectural exploration such as parallelism or reduced/different numerical representation (e.g., Posits~\cite{podobas2018hardware}).

The StreamBrain interface (or language) is heavily inspired by Keras~\cite{ketkar2017introduction}, where the user constructs the network layer-by-layer after finally calling the training function. Currently, we primarily focus on three-layer (input$\rightarrow$hidden$\rightarrow$classification) layer networks. Furthermore, the StreamBrain framework includes data-loaders for several well-known datasets, including MNIST, STL-10, CIFAR10/100, and – more recently – the Higgs dataset~\cite{baldi2014searching}.

\subsection{In-Situ Visualization of BCPNN}
A new StreamBrain feature, which is first introduced in this paper, is the \textit{in-situ} visualization of the training loop during execution.
Often, in machine learning contexts, the training of the network is launched, and the only real metric visible to the user is the loss function that is being optimized, often on a per-epoch resolution. While this works very well in machine learning paradigms that are supervised, in methods that are unsupervised (such as BCPNN), it is often desirable to inspect parameters and variables that gradually grow in an unsupervised manner. More specifically, it is desirable to visually inspect the receptive fields of learnable units in the network.

In BCPNN, HCUs have a trainable receptive field-- each HCUs can learn (through a technique called structural plasticity) wherein the dataset should “look at” to extract a majority of information. This receptive field is trained in an unsupervised and usually, it is updated once per epoch. During the update, the BCPNN structural plasticity (see~\cite{ravichandran2020learning} for a detailed description) will be invoked, which essentially tries to exchange \textit{active} (used) connections with low-entropy for silent (inactive) high-entropy connections. The receptive field properties (e.g., size, as will be seen in the later result section) is very important to obtain good performance.

\begin{figure}
\begin{center}
\includegraphics[width=0.5\textwidth]{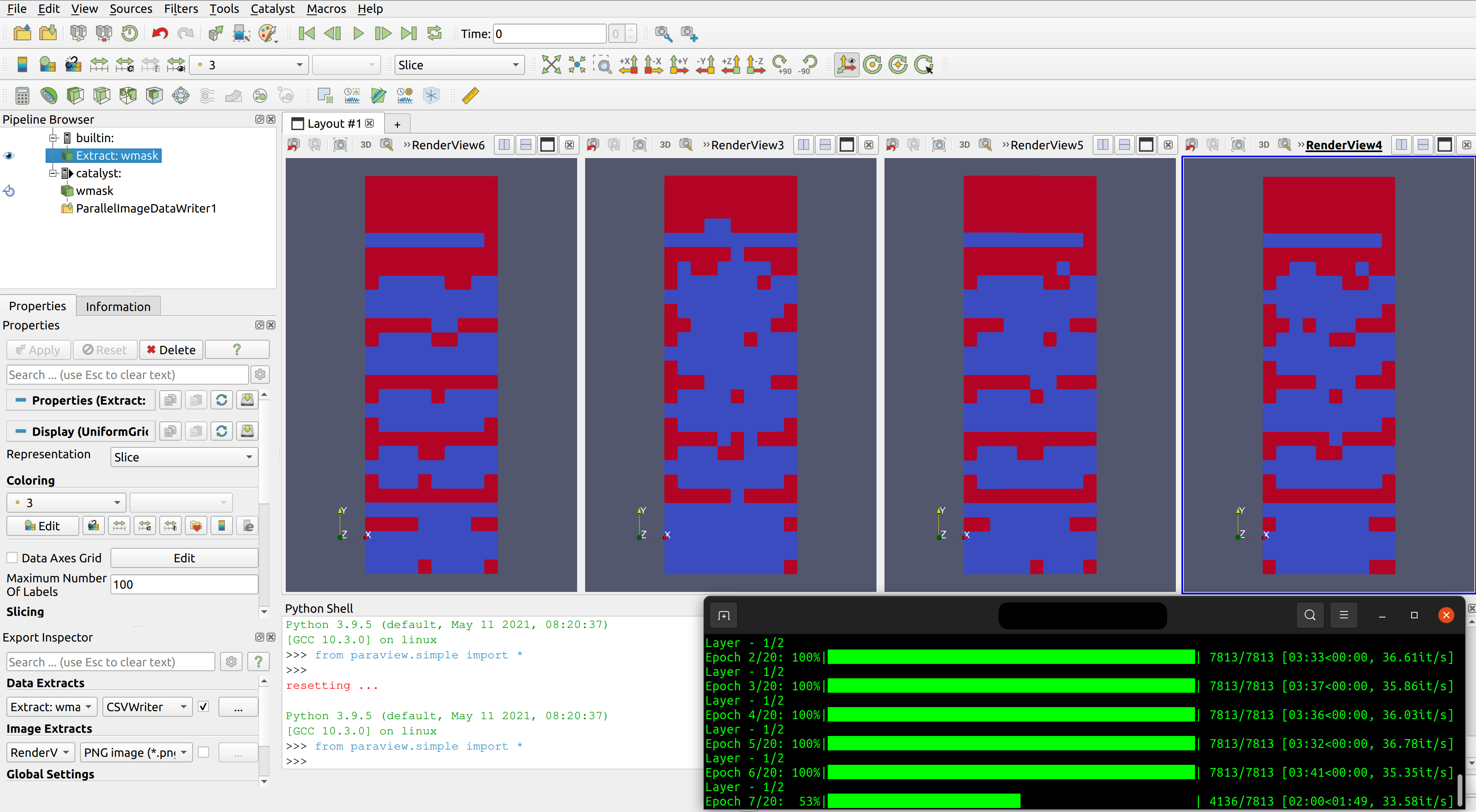}
\end{center}
\vspace{-0.5cm}
\caption{\textit{In-situ} observation and visualization of the development of the receptive fields of the HCUs after training for a number of epochs (red=active connection, blue=silent connection)}
\vspace{-0.5cm}
\label{fig:paraview}
\end{figure}

To facilitate inspection of the receptive fields, we have incorporated prototype support for ParaView Catalyst~\cite{fabian2011paraview}. ParaView is an open-source, multi-platform data analysis and visualization application, which has been used to provide \textit{in-situ} visualization of many different HPC scientific simulations~\cite{atzori2021situ}. In our case, that HPC simulation is the receptive field of the HCUs. We introduce a new StreamBrain visualization module and implemented a Catalyst Adaptor through the Catalyst Python and C++ APIs. The adaptor triggers co-processing at end of each epoch and the Catalyst pipeline writes the receptive fields as VTI files. Furthermore, the ParaView client can accept live connection from Catalyst to, visualize, pause, and inspect the fields as the training progresses. We illustrate in Fig.~\ref{fig:paraview} a training of the Higgs Boson dataset (that will be described in Section~\ref{result} in detail) using four HCUs with a density of 40\%. By looking at ParaView, we can observe how the receptive fields develop over time.

\section{Experimental Set-up}
\label{setup}
We perform our experiments on an A100-system, a local node at KTH with an AMD Epyc 7302P (16-Core) processor, and a powerful NVIDIA Ampere-100 GPU (PCIe)~\cite{svedin2021benchmarking}. We build StreamBrain using GCC 8.3.1, CUDA 11.1, ParaView 5.8, and Python 3.8.5. In this work, we use the fully offloaded CUDA backend for the highest performance.

One major advantage of using a flexible Python framework such as StreamBrain is the ease of exploration. Contrary to traditional DL methods, the formulation of BCPNN implies a larger number of hyperparameters that are use-case-dependent. For this reason, we use the Adaptive Exploration Platform\footnote{\url{https://ax.dev}} together with Nevergrad\footnote{\url{https://github.com/facebookresearch/nevergrad}} to search for optimal hyperparameter combinations. 

\section{A Use Case for StreamBrain: Analysing the Higgs Boson Dataset}
\label{result}
To showcase the StreamBrain capabilities, we analyze the Higgs Boson dataset\footnote{Available at \url{https://archive.ics.uci.edu/ml/datasets/HIGGS}}. The dataset comprises 11 million simulated collisions and is approximately 2 GB large. Each collision process is labeled with either \emph{s} for \emph{signal} or \emph{b} for \emph{background} and consists of 28 input features. We use the Higgs boson dataset to discriminate between a \emph{signal} collision process (a new theoretical Higgs boson is obtained), and a \emph{background} collision process that is not relevant for the identification of the Higgs boson. The input features comprise 21 low-level features and seven high-level features that are derived from the other low-level features. A detailed description of the input dataset is provided in Ref.~\cite{baldi2014searching}. We extract a balanced subset of the training set, and then we compute the 10-quantiles and split the distribution into ten groups with approximately even sizes. The features are then encoded as a one-hot vector of size ten, with the component being hot indicating which quantile the feature belongs to.

\subsection{Experiment \#1: Hyper-Columns and Mini-Column Units}
We start our exploration by looking at the impact of changing the fundamental trainable units in BCPNN: the HCUs and MCUs. More specifically, we are interested in observing the testing performance obtained when increasing the capacity of the network. The capacity of the network, broadly speaking, is dictated by the \# MCUs, since each MCU can capture a particular instance of a variable. Broadly speaking, one MCU in StreamBrain corresponds to a neuron in a traditional neural network. The second hyperparameter to dictate capacity is the \# HCUs. Increasing the number of HCUs increases the number of discrete variables that are modeled, and – unless the receptive field is set to 100\% – increasing the number of HCU will likely provide more extensive coverage of the input data.

To illustrate the interplay between network capacity, testing accuracy, and execution time, we set up three experiments: one experiment with 30 MCUs per HCU, one with 300 MCUs per HCU, and one with 3000 MCUs per HCU. In each experiment, the HCU receptive field is set to 30\%-- in short, each HCU is allowed to have 30\% active connections to the input data, forcing the HCUs to be selective. Next, for each of the above experiments, we vary the number of HCU made available to the network-- the size of the network thus grows linearly with the number of HCUs. We train each experiment 10 times and take the average as a representative value both for training accuracy and train time (seconds).

\begin{figure}[t]
\begin{center}
\includegraphics[width=0.5\textwidth]{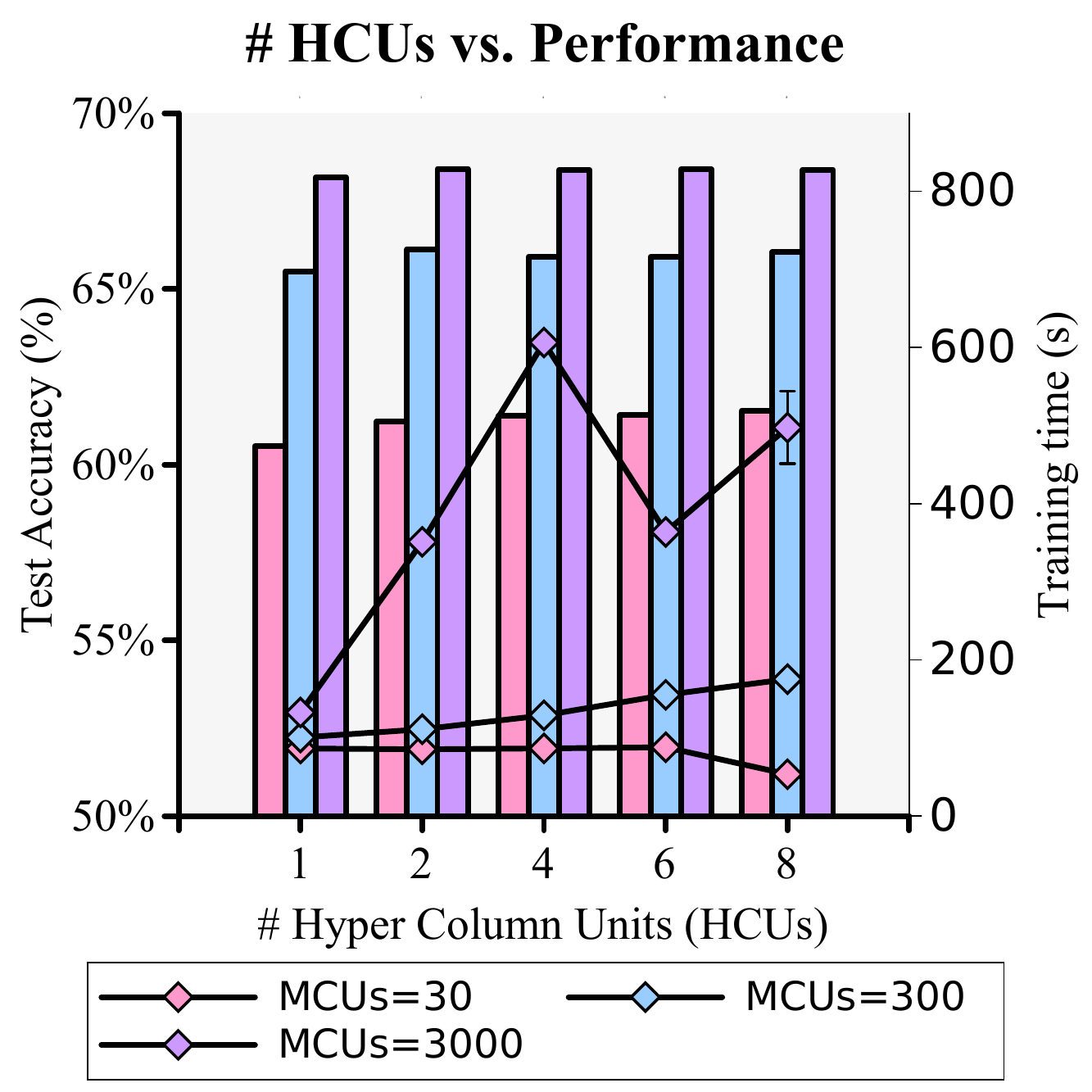}
\end{center}
\vspace{-0.5cm}
\caption{The capacity of the network is dictated by the \# HCUs and \# MCUs, and the larger networks often have a better train accuracy better than smaller one (bars associated with left y-axis). Performance (lines associated with right y-axis) is, as expected, a function of the network capacity.}
\vspace{-0.5cm}
\label{fig:result1}
\end{figure}

The results are shown in Fig.~\ref{fig:result1}. Overall – and in line with expectations – using a single HCU will have the performance primarily dictated by the capacity inside that HCU. For example, going from an HCU that has 30 MCUs to one that has ten times larger capacity (300 HCUs) increases network performance by ~5\%, albeit going another magnitude more in capacity yields only 0.54\% increase. In other words, higher capacity gives higher performance. For the particle dataset that we are considering, increasing the number of HCUs does not seem to yield a noticeable increase in performance. We see a ~0.9\% improvement on the 30 MCUs case when increasing the number of HCUs, but only a ~0.5\% and ~0.3\% improvement for the 300 and 3000 MCUs cases, respectively. The reason behind this is not fully understood yet, and we will further scrutinize this in future work. The training time of the network, however, is a direct function of both the number of MCUs and then the number of HCUs, where training a larger network also consumes more time. In this experiment, the shortest training time was for the smallest network (1 HCU, 300 MCUs) with 86.6 seconds, while the longest training time was consumed by the network with 4 HCUs and 3000 MCUs, taking 606 seconds to train. Interestingly, going beyond 4 HCUs seem to decrease training time considerably, which could be because of the more favorable dimensions for the GPU~\footnote{similar “Jiggs” in performance can be seen in BLAS performance charts for GPU, where some matrix dimension is more favorable than others} but will require more investigation in the future. The largest network experience quite a large standard deviation of ~9.3\%. Combining unsupervised learning in StreamBrain with SGD reaches \textbf{69.15\%} performance on the single HCU (3000 MCU) configuration (with the AUC performance being 76.4\%).

\subsection{Experiment \#2: Magnitude of Receptive Field}
One of the outcomes of exploring the capacity of the network in the previous section was, once the performance saturated when increasing the number of MCUs for a given receptive field, we could increase the receptive field by increasing the number of HCUs (where the receptive field grew linearly with the \# HCUs). Another way of doing this is to increase the size of the receptive field, which is another hyperparameter to tune.

Increase the receptive field can be a good way of increasing the performance of the network since the HCUs now have access to more data. However, at the same time, blindly increasing the receptive field can yield less insight into the data. For example, while it may be tempting to create a network with a single HCU with a 100\% receptive field post-training, little knowledge will be gained about the data stream itself. Hence, from a data-science side, it is crucial to select a receptive field that learns both how to perceive the data and its structure.

\begin{figure}
\begin{center}
\includegraphics[width=0.5\textwidth]{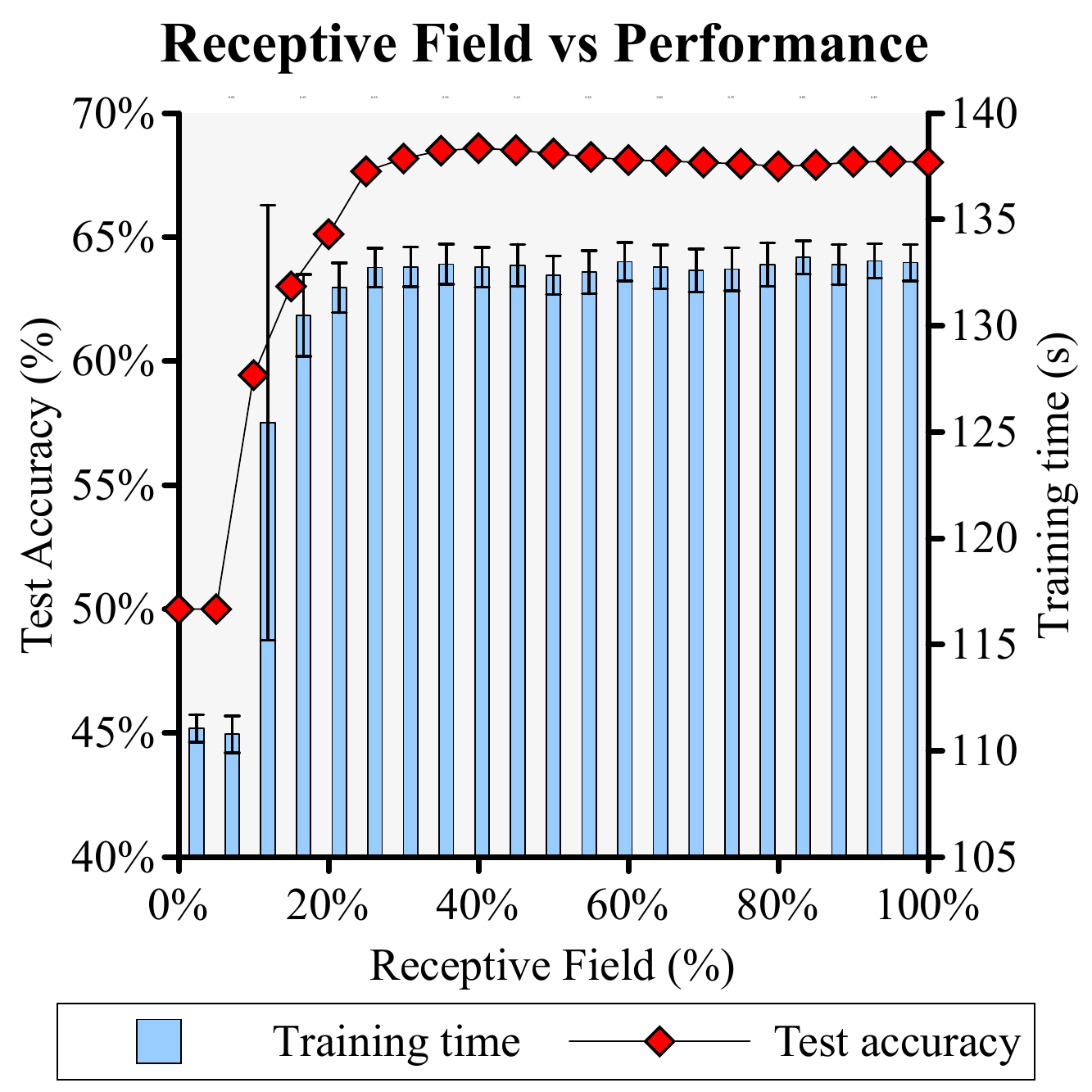}
\end{center}
\vspace{-0.5cm}
\caption{The performance of the network often increases with a larger receptive field (red line, left y-axis). In this example, a 40\% receptive field yields the highest test accuracy of 68.58\%. The time it took to train the network are the blue bars associated with the right y-axis.}
\vspace{-0.5cm}
\label{fig:result2}
\end{figure}

To illustrate the impact of the size of the receptive field as a hyperparameter, we set up an experiment where we fix the network capacity with a configuration consisting of a single HCU with 3000 MCUs (identical to one of our starting points in the previous section) and instead change the receptive field from covering between 0\% and 100\% of the data stream. We executed this experiment 10 times, taking the average as a representative case.

Fig.~\ref{fig:result2} shows our results. The first thing we notice is that using a very small receptive field between 0\%-5\% limits the trainable performance to mere chance (50\%), and it is not before we reach a 10\% receptive field that we start to climb in training accuracy. The training accuracy increases with a larger receptive field and reaches a climax at 68.58\% when the HCU is allowed to look at nearly half (40\%) of the data stream, after which no improvement can be seen-- increasing the receptive field further beyond this point does not positively impact the training accuracy. At this point, the AUC performance is 75.5\%.

Unlike changing the capacity of the network (previous section), changing the size of the receptive influence the execution time only very little, where having a near 0\% mask takes on average 111 seconds to train while having a 100\% mask takes 132.9 seconds to train. This is in line with expectations, as the computation required to train is independent of the size of the receptive field, and only the structural plasticity (which is quite rarely updated) is affected.

\begin{figure}
\begin{center}
\includegraphics[trim={5cm 0.4cm 0 0},clip, width=0.5\textwidth]{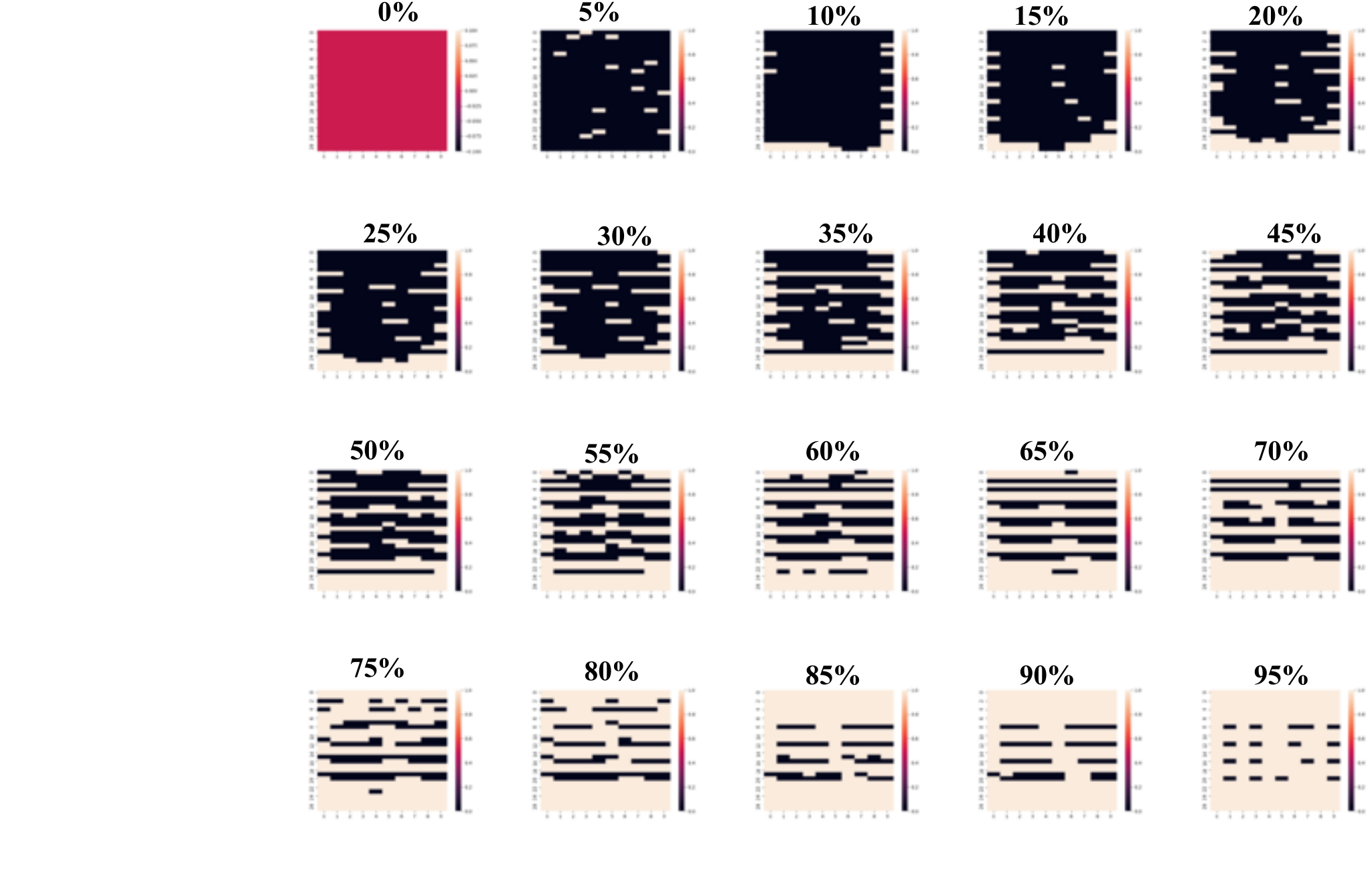}
\end{center}
\vspace{-0.5cm}
\caption{Evolution of masks when increasing the size of the receptive field, starting from 0\% (or zero visibility, top-left) to 95\% (bottom-right).}
\vspace{-0.5cm}
\label{fig:result3}
\end{figure}

The masks and their evolution can be seen in Fig.~\ref{fig:result3}, where we have picked a subset of the masks produced in the experiment, and we can see that as we increase the size of the receptive field, the HCU covers a larger and larger area of the image. Note that depending on the size of the receptive field, the BCPNN framework does not necessarily pick the same connections, but the best connections for a (e.g.,) a 5\% receptive field not necessarily be included in (e.g.,) a 10\% receptive field.

\section{Related Work}
\label{related}
BCPNN -- which StreamBrain implements -- has been used in prior work to study both MNIST~\cite{ravichandran2020learning} and STL-10~\cite{podobas2021streambrain} networks. The present study is the first time BCPNN has been applied in a  dataset associated with physics. However, several other studies have been done on similar datasets. For example, there was a well-known Kaggle challenge~\footnote{https://www.kaggle.com/c/higgs-boson} on using the results from the Atlas experiment to find new particles. In 2014 – when the challenge ended – the winning network was a bag of 70 dropout neural networks that were created through cross-validations on the data. The Kaggle challenge was not scores based on classification, but rather on a different metric (approximate median significance, AMS). 
The data-set that we used was explored with other machine learning strategies, including Boosted Decision Trees,  Shallow Neural Networks, and Deep Neural Networks. They reach between 81.6\% (MLP) and 88\% (Deep Neural Network) of Area Under the Curve (AUC) performance, while we reach 75.5\% (only BCPNN with 1 HCU) and 76.4\% (by combining BCPNN+SGD).


\section{Discussion and Conclusions}
\label{conclusion}
In this paper, we have described our experience in applying the BCPNN ML method through the StreamBrain framework on a high-energy particle physics dataset. We have shown how StreamBrain's unique characteristics can help gain insight into the data stream through inspection of receptive fields. Furthermore, we have quantified the training time associated with using the StreamBrain framework on a state-of-the-art GPU. This early work reaches 68.5\% (or 69.15\% by mixing BCPNN with SGD) test accuracy on the data set, which -- while being lower than using existing methods -- encourages us to further develop the framework for use in particle physics. Among the future direction is to use more HCUs and hybrid training to improve the quality of the solution even more, as well as adapting hyperparameters associated with structural plasticity dynamically online, possibly guided by an end-user through the ParaView visualization.

\noindent{{\textbf {\footnotesize Acknowledgements: }}\scriptsize We acknowledge funding from the European Commission H2020 program, Grant Agreement No. 801039 (EPiGRAM-HS), and No. 800999 (SAGE2).}
\vspace{-0.3cm}


\bibliographystyle{IEEEtran}
\bibliography{main}
\end{document}